\documentclass[10pt,conference]{IEEEtran}
\IEEEoverridecommandlockouts % show footnote comments
\usepackage{algpseudocode}                                 % new algorithm package

\usepackage{algorithm}
\usepackage{graphicx}                                      % include pdf figures
\usepackage{amsmath}
\usepackage{amssymb}
\usepackage{amsfonts}
\usepackage{amsthm}
\usepackage[mathcal]{eucal}
\usepackage{booktabs}
\usepackage{enumerate}
\usepackage{multirow}
\usepackage[subrefformat=parens,farskip=0pt,justification=centering]{subfig}
\usepackage{color}
\usepackage{cite}                                          % more citations in one bracket
\usepackage{comment}                                       % use comment
\usepackage{soul}                                          % use highlight command \hl{}
\soulregister\cite7
\soulregister\ref7
\soulregister\pageref7
\usepackage{etoolbox}                                      % commands \newtoggle, \toggletrue, \iftoggle
\usepackage{url}
\usepackage{nth}                                           % nth command
\usepackage{bm}                                            % bm command
\usepackage{courier}
\usepackage{balance}
\usepackage{threeparttable}
\usepackage[bookmarks=false]{hyperref}
\hypersetup{
    colorlinks = true,
    citecolor  = blue,
    linkcolor  = blue,
    urlcolor   = blue,
}
\usepackage{tikz}
\usetikzlibrary{positioning,calc,fit,decorations.pathmorphing}
\usepackage{filecontents}                                  % support to pgfplots
\usepackage{pgfplots}
\usepackage{pgfplotstable}
\pgfplotsset{compat=newest}
\usepackage{caption}
\usepackage{cleveref}
\Crefformat{figure}{Fig.~#2#1#3}                           % "Fig.", instead of "Figure"
\Crefname{subfigure}{Fig.}{Figs.}
\Crefname{figure}{Fig.}{Figs.}
\Crefformat{table}{TABLE~#2#1#3}                           % "TABLE", instead of "Table"
\definecolor{CUHKorange}{RGB}{244,106,18} %F47012
\definecolor{CUHKblue}{RGB}{0,111,190}    %006FBE
\definecolor{CUHKgreen}{RGB}{0,127,128}   %007F80
\definecolor{CUHKred}{RGB}{228,46,36}     %E42E24
\definecolor{CUHKyellow}{RGB}{198,148,34} %C69422
\definecolor{CUHKdark}{RGB}{114,44,114}   %722C72
\definecolor{CUHKmiddle}{RGB}{144,44,144} %902C90
\definecolor{CUHKlight}{RGB}{167,44,167} 

\renewcommand{\vec}[1]{\boldsymbol{#1}}    % re-define vec command

\newtheorem{myassumption}{Assumption}

\algrenewcommand\textproc{\texttt}

\makeatletter
\let\OldStatex\Statex
\renewcommand{\Statex}[1][3]{%
  \setlength\@tempdima{\algorithmicindent}%
  \OldStatex\hskip\dimexpr#1\@tempdima\relax
}
\makeatother

\RequirePackage[normalem]{ulem} %DIF PREAMBLE
\RequirePackage{color}\definecolor{RED}{rgb}{1,0,0}\definecolor{BLUE}{rgb}{0,0,1} %DIF PREAMBLE
\graphicspath{{./figs/}}

\definecolor{myorange}{RGB}{238,97,42}  %
\definecolor{myblue}{RGB}{178,179,249}  %A2A0FE
\definecolor{mygrey}{RGB}{166,166,166}  %
\definecolor{mygreen}{RGB}{180,210,36}  %B4D224
\definecolor{myred}{RGB}{238,0,0}       %EE0000
\definecolor{myyellow}{RGB}{198,148,34} %C69422
\definecolor{mydark}{RGB}{114,44,114}   %722C72
\definecolor{mymiddle}{RGB}{144,44,144} %902C90
\definecolor{mylight}{RGB}{167,44,167}  %A72CA7

\title{Routing Towards Discriminative Power of \\ Class Capsules}

\author{
    Haoyu Yang, \quad
    Shuhe Li, \quad
    Bei Yu \\
    Department of Computer Science and Engineering \\
    The Chinese University of Hong Kong 
}

\begin{document}

\maketitle
\begin{abstract}
Capsule networks are recently proposed as an alternative to modern neural network architectures.
Neurons are replaced with capsule units that represent specific features or entities with normalized vectors or matrices.
The activation of lower layer capsules affects the behavior of the following capsules via routing links that are constructed during training via certain routing algorithms.
We discuss the routing-by-agreement scheme in dynamic routing algorithm which, in certain cases, leads the networks away from optimality.
To obtain better and faster convergence, we propose a routing algorithm that incorporates a regularized quadratic programming problem which can be solved efficiently.
Particularly, the proposed routing algorithm targets directly on the discriminative power of class capsules making the correct decision on input instances.
We conduct experiments on MNIST, MNIST-Fashion, and CIFAR-10 and show competitive classification results compared to existing capsule networks.
\end{abstract}

\section{Introduction}
\label{sec:intro}

Convolutional neural networks have been deeply studied in recent years.
Its variations are successfully and widely applied in different tasks including classification \cite{DL_NIPS2012_Krizhevsky}, generation \cite{DL_NIPS2014_Ian}, segmentation \cite{DL_CVPR2015_Long}, and so on.
Convolution layers abstract common features hierarchically by scanning the object with shared kernels
that decomposes the original images into small and simple instances which are hence used for classification task.
However, the process to some extent violates the nature of recognizing objects, that the visual system resembles parse tree-like \cite{CAP_NIPS2000_Hinton} structures on fixation points adopted by human vision.
In each layer of a parse tree, neurons are grouped together representing certain objects, which are known as capsules.

\begin{figure}[tb!]
	\centering
	\includegraphics[width=1.0\linewidth]{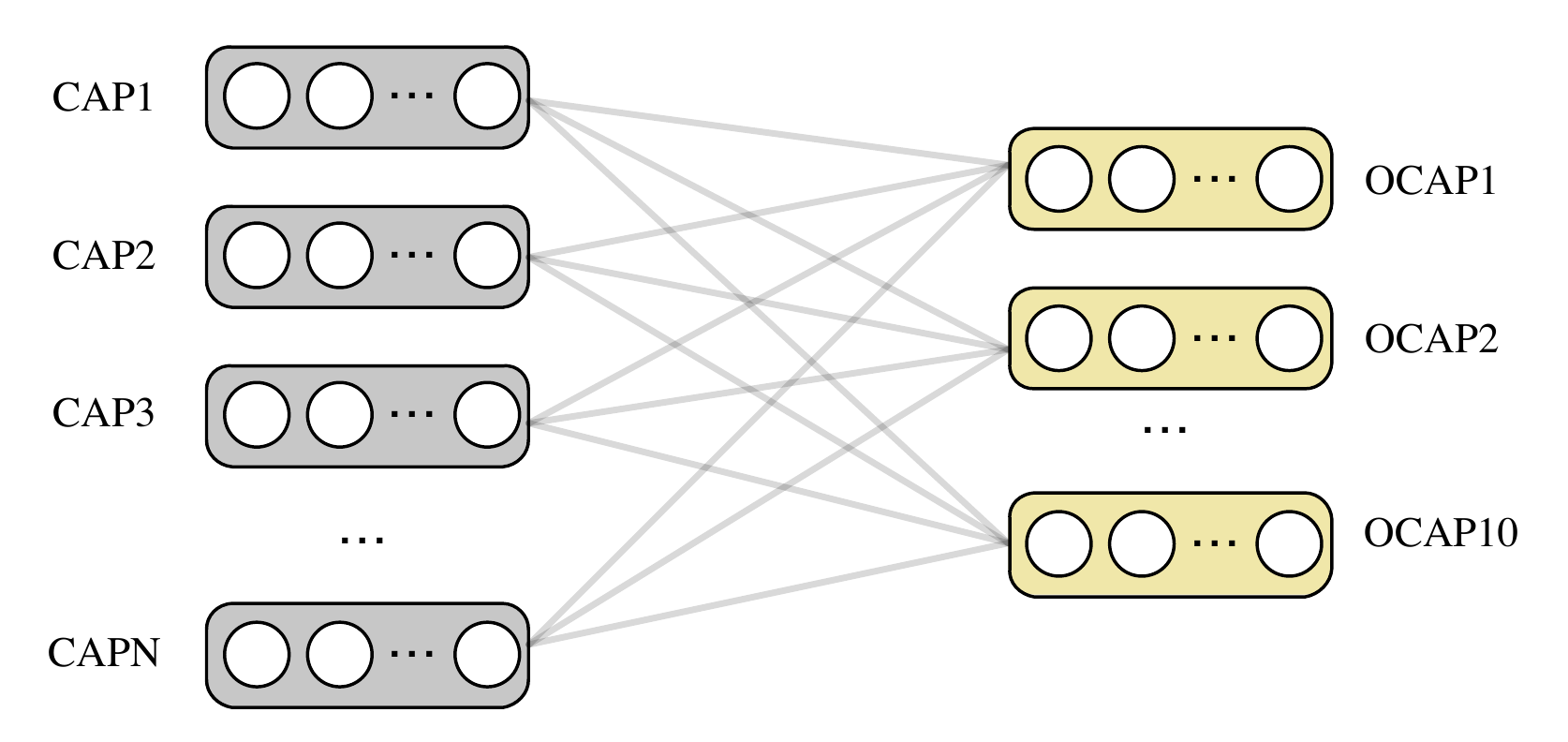}
	\caption{Visualization of capsule layers with 10 capsules in the output layer that represent the existence of 10 classes.}
	\label{fig:capnet}
\end{figure}

Capsule networks are recently proposed as an alternative of modern convolutional neural network architectures that changes the way neural networks are trained and features are represented,
and, as a result, it also brings robustness to adversarial attacks and overlapped objects \cite{CAP_NIPS2017_Hinton}.
A capsule is a group of neurons that represent a feature or an entity. 
The capsule length reflects by how much the capsule is activated or the probability a corresponding entity exists in a given image.
Capsules in adjacent layers are densely connected via traditional neuron links with their weights learned through routing-by-agreement algorithms, as shown in Figure~\ref{fig:capnet}.
Another characteristic of capsule networks is that lower level features are constructing higher level entities as layer goes deeper,
compared to convolutional neural networks that perform feature abstraction layer by layer.

Two recent capsule routing algorithms are dynamic routing \cite{CAP_NIPS2017_Hinton} and EM routing \cite{CAP_ICLR2018_Hinton}.
Dynamic routing quantifies the agreement between capsules via their inner-product.
The greater inner-product value indicates two capsules agree more with each other and the dynamic routing aims to amplify the agreement.
EM routing models each higher level capsule as a Gaussian and the posterior probability of previous layer capsules determines in which level they are connected to higher level capsules.
In both routing algorithms, capsules are coupled to higher level capsules according to certain agreement metrics without considering the prediction results.
%In dynamic routing, for example, a primary capsule will have larger inner-product with a digital capsule if they already have the largest inner-product among all the primary capsules,

In this work, we discuss and analyze the routing-by-agreement mechanisms in capsule networks and propose a routing algorithm
that can achieve faster convergence and better discrimination results in classification tasks.
The algorithm is inspired by two observations: (1) the ultimate objective of training capsule networks is to make the correctly activated output capsules have the largest lengths and 
(2) the feature capsules (capsules before the output layer) are reasonable to have negative effects on capsules in the following layers.
We also propose several training tricks to enlarge the solution space that can result in higher classification accuracy on several datasets.
We pick the capsule network architecture used in \cite{CAP_NIPS2017_Hinton} as a case study to show how our methods benefit the training of capsule networks.
\iffalse
The main contributions of this paper are summarized as follows:
\begin{enumerate}
	\item We revisit the dynamic routing algorithm in \cite{CAP_NIPS2017_Hinton} and discuss two beneath assumptions in the algorithm, which might not be necessarily hold in some cases.
	\item We establish a new objective on determining routing coefficients between capsules in adjacent capsule layers according to the classification mechanism of capsule networks.
	\item Two popular regularization methods are discussed and embedded in our routing algorithm that can be solved efficiently within the neural network training procedure.
	\item We conduct experiments on three widely used benchmarks and compare the results with dynamic routing algorithm which shows the effectiveness of the proposed methods.
\end{enumerate}
\fi

\section{Related Works}
\label{sec:prelim}
\cite{LEARN_NIPS2000_Hinton} developed credibility networks where images are interpreted as a parse tree or a collection of parse trees with the leaf nodes being image pixels.
Each node and its associated latent variables represents an object and its pose information that forms higher level objects.
The concept of capsule comes from transforming auto-encoders \cite{CAP_ICANN2011_Hinton}, where each capsule consists of an instantiation of certain entity.
All the entities together reconstructs images with some transformation that is applied on the instantiation vectors with some probabilities.
Although the capsule in \cite{CAP_ICANN2011_Hinton} aims at reconstruction, it already comes with similar features of the capsule discussed in this paper, i.e., associating with a higher level object in a feedforward style.
\cite{CAP_NIPS2017_Hinton} and \cite{CAP_ICLR2018_Hinton} are two latest implementations of capsule networks for object classification, certain routing algorithms have been discussed in previous section.
\cite{CAP_ICLR2018_Wang} rephrased the dynamic routing algorithm as a KL divergence regularized clustering problem that inspires
an improved solution resembling agglomerative fuzzy k-means, which can be solved by coordinate descent.

\section{The Algorithm}
\label{sec:algo}

In this section, we will discuss the details of the routing-by-agreement scheme in the dynamic routing algorithm
based on which, an improved routing algorithm is proposed that targets directly at the discriminative power of class capsules.

\subsection{Dynamic Routing-by-Agreement}
Routing-by-agreement aims to couple the lower level capsules to higher level capsules when they agree with each other.
Here we will discuss the coupling procedure from primary capsules to output capsules in \cite{CAP_NIPS2017_Hinton}.
Each primary capsule $\vec{u}_i$ is first projected to the space of digital capsules in the follow-up layer by %$\hat{\vec{u}}_{j|i}=\vec{W}_{ij} \vec{u}_i$
\begin{align}
	\label{eq:p2d}
	\hat{\vec{u}}_{j|i}=\vec{W}_{ij} \vec{u}_i,
\end{align}
and the digital capsules are then derived from the weighted summation of all $\hat{\vec{u}}_{j|i}$s, as in \Cref{eq:dcap},
\begin{align}
	\label{eq:dcap}
	\vec{v}_j = \sum_{i} c_{ij}\hat{\vec{u}}_{j|i},\vec{s}_j = \text{squash}(\vec{v}_j).
\end{align}
where $\text{squash}$ brings nonlinearity to digital capsules and scales capsule length to between 0 and 1,
\begin{align}
	\label{eq:squash}
	\text{squash}(\vec{v}) = \dfrac{||\vec{v}||^2_2}{1+||\vec{v}||^2_2} \cdot \dfrac{\vec{v}}{||\vec{v}||_2},
\end{align}
and $c_{ij}$s are softmaxed coupling coefficients $b_{ij}$s (see \Cref{eq:rsftmx}) that determines the probability on primary capsule $\vec{u}_i$ should contribute to activate the digital capsule $\vec{v}_j$.
\begin{align}
	\label{eq:rsftmx}
	c_{ij}=\dfrac{\exp(b_{ij})}{\sum_{k} \exp(b_{ik})}.
\end{align}
In each routing iteration, $c_{ij}$ will be amplified if capsule $i$ agrees with capsule $j$ the most.
There are two beneath assumptions in the dynamic routing algorithm.
\begin{myassumption}
\label{asm:1}
Primary capsules do not have negative impact on the activation of digital capsules.
\end{myassumption} 
\begin{myassumption}
\label{asm:2}
All digital capsules are activated correctly.
\end{myassumption} 

Assumption~\ref{asm:1} comes from the fact that $c_{ij}$s are always positive due to \Cref{eq:rsftmx}, which guarantees each primary capsule will more or less contribute to higher level capsules.
Such design cannot efficiently represent the case when one or more specific entities/features can never exist in certain objects.
The dynamic routing algorithm is coherent with Assumption~\ref{asm:2}, because in each routing iteration, $c_{ij}$ will always be increased if $\hat{\vec{u}}_{j|i}$ has largest inner product with $\vec{v}_j$.
Here are some potential drawbacks when accepting these assumptions. 
Assumption~\ref{asm:1} can possibly limit the solution space with always-nonnegative $c_{ij}$'s.
On the other hand, more importantly, Assumption~\ref{asm:2} does not necessarily hold during training especially at very early epochs.
Unconditionally coupling with a digital capsule simply based on inner-product agreement even that capsule is incorrectly activated will hold back the whole training procedure.
According to the observations above, we will introduce an improved routing algorithm that is expected to achieve better classification results and faster convergence.

\subsection{Routing Towards Discriminative Quality}
The length of a capsule is originally designed as indicators of the existence of corresponding features.
Features with larger capsule vector length are more likely to exist in a given instance.
For simplicity, we will use the architecture in \cite{CAP_NIPS2017_Hinton} in the following discussion.
Digital capsules, as the output layer of capsule networks, make the final decision of prediction tasks.  
Therefore, the activation error should be considered when determining routing coefficients.
According to the prediction mechanism of capsule networks, intuitively, the length of capsules that are supposed to be activated should be maximized,
while the length of inactivated capsules should be minimized.
This objective can be written in a unified form as shown in \Cref{eq:obj-1}.
\begin{subequations}
\label{eq:obj-1}
\begin{align}
\max_{\vec{b}_j}~~ &\delta_{ij} ||\vec{v}_j||_2^2,\\
\mathrm{s.t.~} 	&\delta_{ij}= \begin{cases}
    1, & \text{if } i=j, \\
    -1,& \text{otherwise},
\end{cases}
\end{align}
\end{subequations}
where $i$ corresponds to the labels of given observations and the indicator function $\delta_{ij}$ ensures \Cref{eq:obj-1} consistent with the discrimination mechanism of digital capsules.
We denote $\vec{b}_j$ as the routing coefficients corresponding to the $j^{th}$ output capsule before going into softmax function.
To additionally enlarge the representation space of digital capsules, we also discard the softmax of routing coefficients such that each digital capsule is calculated directly through
\begin{align}
	\vec{v}_j &= \sum_{i} b_{ij}\hat{\vec{u}}_{j|i} =\hat{\vec{U}}^\top\vec{b}_j,
\end{align}
where rows of $\hat{\vec{U}}$ are the primary capsules projected into digital capsule space with $\hat{\vec{u}}_{j|i}$,
and the objective of \Cref{eq:obj-1} becomes
\begin{align}
\label{eq:obj-1-1}
\max_{\vec{b}_j}~~ \delta_{ij,k} \vec{b}_j^\top \hat{\vec{U}}_k\hat{\vec{U}}_k^\top \vec{b}_j.  
\end{align}
Note that \Cref{eq:obj-1-1} can not fit each individual observation in the training dataset, because it will always give optimal optimal solution of
\begin{align}
\vec{b}_j^\ast=
\begin{cases}
    \vec{0}, & \text{if } \delta_{ij,k}=-1, \\
    \inf,       & \text{if } \delta_{ij,k}=1.
\end{cases}
\end{align}
Rewrite \Cref{eq:obj-1-1} into batch mode we have a slightly better formulation:
\begin{align}
	\label{eq:obj-1-b}
	\max_{\vec{b}_j}~~ \sum_{k} \delta_{ij,k} \vec{b}_j^\top \hat{\vec{U}}_k\hat{\vec{U}}_k^\top \vec{b}_j,
\end{align}
where $j$ corresponds to the $j^{th}$ digital capsule and $k$ is the $k^{th}$ observation in the training batch.
We are able to obtain a local optimal of \Cref{eq:obj-1-b} as long as $\sum_{k}\delta_{ij,k} \hat{\vec{U}}_k\hat{\vec{U}}_k^\top\nsucceq \vec{0}$. 

\subsubsection{$l_2$-Regularization}
Observing that each capsule is equipped with very small number of neuron nodes that makes $\hat{\vec{U}}_k \in \mathbb{R}^{m\times n}$ have very few columns and as a result, $\hat{\vec{U}}_k\hat{\vec{U}}_k^\top$ has a very low rank
that also makes it possible to turn \Cref{eq:obj-1-b} into a ridge regression-like \cite{LEARN_TM1970_Hoerl} problem with a small regularization on $\vec{b}_j$, as in \Cref{eq:obj-2}.
\begin{align}
	\label{eq:obj-2}
	\max_{\vec{b}_j}~~ \sum_{k=1}^p \delta_{ij,k} \vec{b}_j^\top \hat{\vec{U}}_k\hat{\vec{U}}_k^\top \vec{b}_j
	-\lambda ||\vec{b}_j||_2^2, 
\end{align}
where $\lambda$ is the regularization coefficient which is usually set small, e.g.~$0.001$.
Note that \Cref{eq:obj-2} is no longer convex that makes the maximization reasonable, by the fact that
\begin{equation}
    \begin{aligned}
        &\sum_{k=1}^p \delta_{ij,k}(\vec{b}_j^\top \hat{\vec{U}}_k\hat{\vec{U}}_k^\top \vec{b}_j) - \lambda ||\vec{b}_j||_2^2 \\
        ={}&\vec{b}_j^\top (\sum_{k=1}^p \delta_{ij,k}\hat{\vec{U}}_k\hat{\vec{U}}_k^\top -\lambda \vec{I}) \vec{b}_j \\
        ={}&\vec{b}_j^\top \vec{Q}  (\vec{\Lambda} -\lambda \vec{I}) \vec{Q}^\top \vec{b}_j \nsucceq \vec{0},
    \end{aligned}
\end{equation}
where $\lambda>0$ and $\vec{\Lambda}$ is a diagonal matrix with as least $m-pn$ zeros in its diagonal that ensures $\vec{\Lambda} -\lambda \vec{I}$ to be indefinite as long as the batch size $p$ is not extremely large.
The regularization term also avoids $\vec{b}_j$ going too large or too small that resembles momentum in classical neural network training algorithms \cite{LEARN_NN1999_Ning}.
Because the capsule coupling coefficients are designed to approximate a large set of observations,
we solve the problem greedily by ascending the gradient of on a batch of observations, as in \Cref{eq:gaobj2}.
Let
\begin{align}
r=\sum_{k=1}^p \delta_{ij,k}(\vec{b}_j^\top \hat{\vec{U}}_k\hat{\vec{U}}_k^\top \vec{b}_j)
-\lambda ||\vec{b}_j||_2^2,
\end{align}
and $\vec{b}_j$ can be updated as follows:
\begin{align}
	\label{eq:gaobj2}
    \vec{b}_j
    ={}&\vec{b}_j+\gamma \dfrac{\partial r}{\partial \vec{b}_j} \nonumber \\
    ={}&\vec{b}_j+\gamma \sum_{k=1}^p \dfrac{\partial \vec{b}_j^\top (\delta_{ij,k} \hat{\vec{U}}_k\hat{\vec{U}}_k^\top -\dfrac{\lambda}{p} \vec{I}) \vec{b}_j}{\partial \vec{b}_j} \nonumber \\
    ={}&\vec{b}_j+2\gamma \sum_{k=1}^p (\delta_{ij,k} \hat{\vec{U}}_k\hat{\vec{U}}_k^\top -\dfrac{\lambda}{p} \vec{I})\vec{b}_j,
\end{align}
where $p$ denotes the observation batch size.

\subsubsection{$l_1$-Regularization}
In the original capsule networks design, primary capsules and digital capsules are densely connected.
It has been shown in previous works such densely connected structure is easily suffering from overfitting \cite{DL_JMLR2014_Nitish,DL_ARXIV2012_Hinton}.
Enforcement weight sharing in CNN and drop neurons when training densely connected nets (also known as dropout) are two major solutions in deep learning scope.
Weight sharing is similarly applied with the implementation of capsule networks in \cite{CAP_ICLR2018_Hinton}.

Instead of predetermine the neuron/capsule connectivity or randomly drop connection, we propose an alternative that can automatically learn how capsules in different layers are linked with each other.
The routing objectives can be found in \Cref{eq:obj-3},
\begin{align}
\label{eq:obj-3}
\max_{\vec{b}_j}~~ \sum_{k=1}^p \delta_{ij,k} \vec{b}_j^\top \hat{\vec{U}}_k\hat{\vec{U}}_k^\top \vec{b}_j
-\lambda ||\vec{b}_j||_1, 
\end{align}
where an $l_1$ penalty term is applied on $\vec{b}_j$ that admits a sparse solution \cite{LEARN_JRSS1996_Robert}.
Because solving \Cref{eq:obj-3} requires to calculate the gradient of $|x|$ at $x=0$, we define	$\left.\frac{\partial |x|}{\partial x}\right \vert_{x=0}=0$.
Routing coefficients can then be similarly updated as follows:
\begin{align}
	\label{eq:gaobj3}
	\vec{b}_j=\vec{b}_j + 2\gamma (\sum_{k=1}^p\vec{b}_j^\top \delta_{ij,k} \hat{\vec{U}}_k\hat{\vec{U}}_k^\top - \lambda \dfrac{\partial ||\vec{b}_j||_1}{\partial \vec{b}_j}).
\end{align}

\subsection{Training Capsule Networks}
Note that for both strategies in \Cref{eq:gaobj2} and \Cref{eq:gaobj3} are compatible with networks that contain more than 2 capsule layers, where the routing coefficients can be accordingly updated through chain rule.
When training other neuron weights, we adopt the margin loss as in \Cref{eq:ml} \cite{CAP_NIPS2017_Hinton}.
\begin{equation}
    \begin{aligned}
        \label{eq:ml}
        L_k ={}& T_k \max(0, m^{+}-||\vec{v}_k||_2)^2 \\
        &+ \lambda^\prime (1-T_k) \max(0, ||\vec{v}_k||_2-m^{-})^2,
    \end{aligned}
\end{equation}
where $T_k=1$ if class $k$ is present in the $k^{th}$ output capsule.

As shown in \Cref{alg:trcap}, the routing coefficients and other neuron weights are updated alternatively in each training step,
where $n_d$, $n_b$ and $n_r$ are the number of output capsules, the number of iterations to update regular weights and the number of iterations for routing, respectively.
In each training iteration, we first sample a minibatch of observations from the training set (lines 1--2),
we then update the regular neuron weights for $n_b$ steps with routing coefficients fixed (lines 3--6),
and finally the routing coefficients are updated according to the formulation in \Cref{eq:obj-2} or \Cref{eq:obj-3} (lines 7--9).

\begin{algorithm}[h]
	\small
	\caption{\small Training Capsule Networks. 
		Routing coefficients $\vec{b}_j, j=1,2,...,n_d$ and regular neuron weights $\vec{W}$ are updated alternatively.
		In each iteration, $n_r$ steps routing and $n_b$ steps back-propagation are conducted respectively. We pick $n_r=1$ and $n_b=1$ in all the experiments.
		}
	\label{alg:trcap}
	\begin{algorithmic}[1]
	\For{number of training iterations}
	\State Sample a minibatch of $p$ observations \{$\vec{x}_i| i=1,2,...,p$\} from the training dataset;
	\For{$n_b$ steps}
	\State Update $\vec{W}$ by descending its gradient; 
	\State $\vec{W} \gets \vec{W}-\dfrac{1}{p} \sum_{i=1}^{p} \sum_{j=1}^{n_d} \dfrac{\partial L_k}{\partial \vec{W}}$;
	\EndFor
	\For{$n_r$ steps}
	\State Update $\vec{b}_j$s by ascending its gradient as in \Cref{eq:gaobj2} or \Cref{eq:gaobj3};
	\EndFor
	\EndFor
	\end{algorithmic}
\end{algorithm}

\section{Experiments}
\label{sec:exp}

\begin{table*}[tb!]
	\caption{Neural network configuration for each benchmark.}
	\label{tab:arch}
	\renewcommand{\arraystretch}{1.3}
	\setlength{\tabcolsep}{1pt}
	\centering
	\small
	\begin{tabular}{c|c|c|ccc}
		\toprule
		\multirow{2}{*}{Layer} & \multirow{2}{*}{Filter/Capsule Size} & \multirow{2}{*}{Activation} & \multicolumn{3}{c}{Filter/Capsule/Neuron Number} \\ 
		&                      &                                      & MNIST                       &Fashion-MNIST             & CIFAR-10                   \\ \midrule
		Conv1                  & 9$\times$9                           & ReLU                        & 256     & 256                 & 256                     \\
		Cap1                   & 8                                    & Squash                      & 32      & 32                   & 64                      \\
		Cap2                   & 16                                   & Squash                      & 10       & 10                 & 10                      \\
		FC1                    & -                                    & ReLU                        & 512     & 512                 & -                       \\
		FC2                    & -                                    & ReLU                        & 1024      & 1024                & -                       \\
		FC3                    & -                                    & Sigmoid                     & 784     & 784                     & -                       \\ \bottomrule
	\end{tabular}
\end{table*}

\begin{figure*}[bt!]
	\centering
	\subfloat[input]{\includegraphics[width=.25\textwidth]{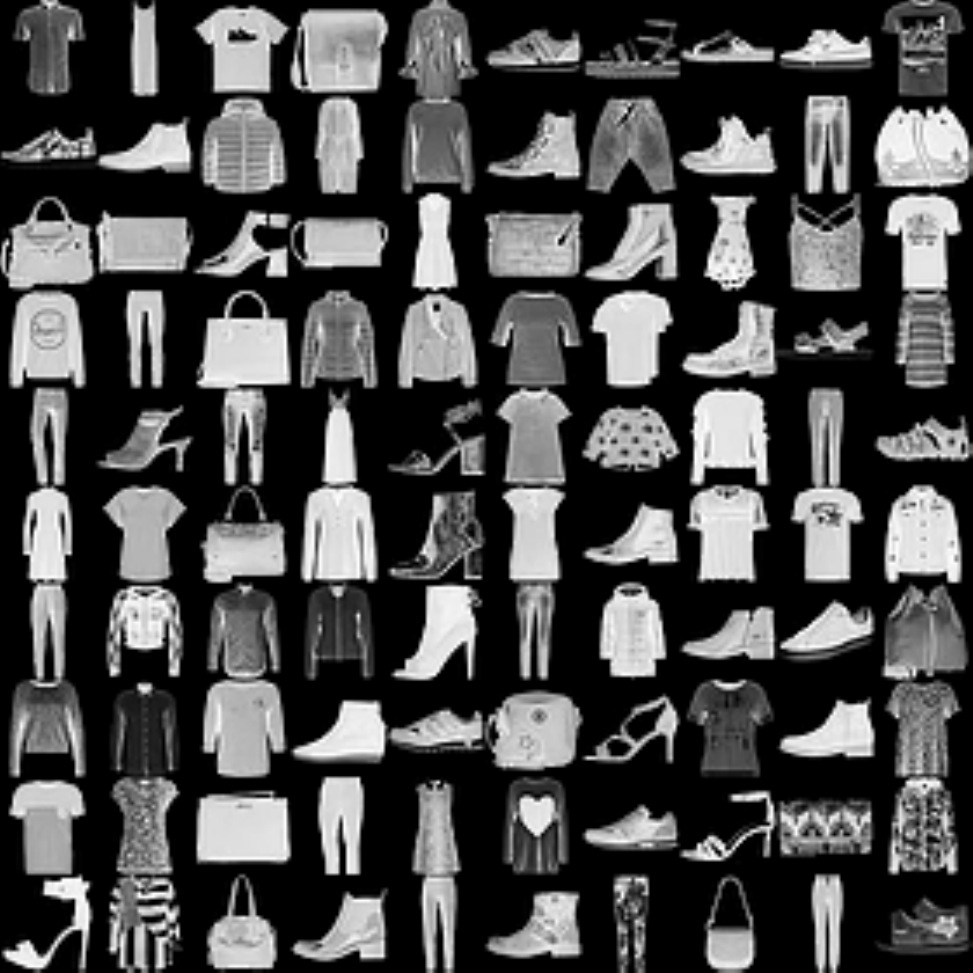}}
	\hspace{0.5cm}
	\subfloat[dynamic routing]{\includegraphics[width=.25\textwidth]{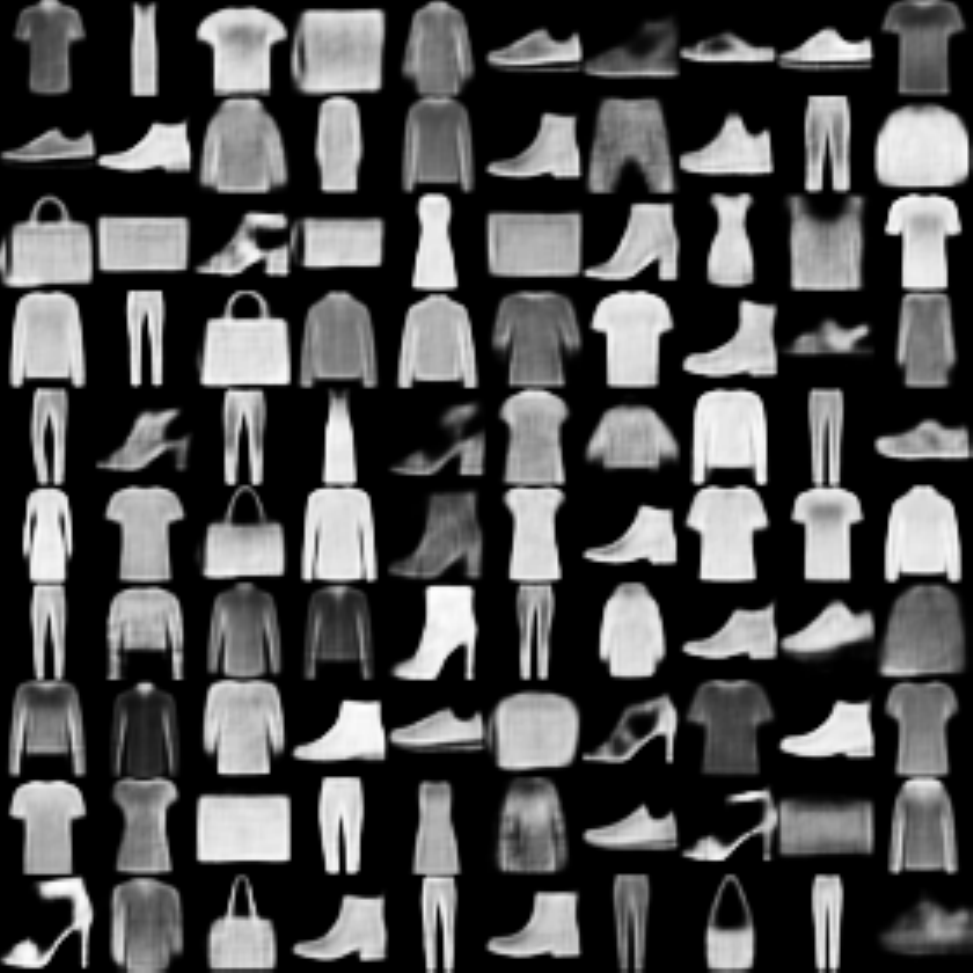}}
	\hspace{0.5cm}
	\subfloat[$l_2$-regularized routing]{\includegraphics[width=.25\textwidth]{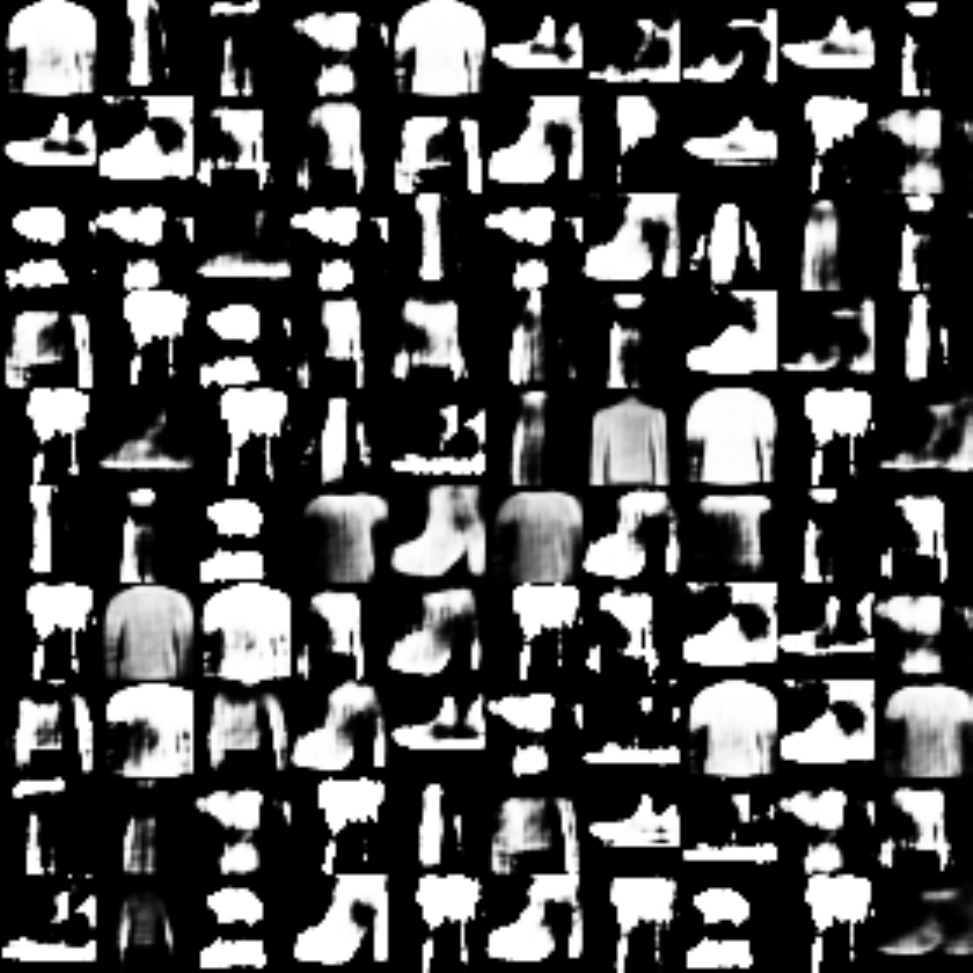}}
	\caption{Visualization of the reconstructed images on Fashion-MNIST dataset.
	(a) 100 image samples from the Fashion-MNIST dataset that can be correctly classified by the capsule networks that is trained with our algorithm;
	(b) The corresponding images reconstructed from the capsule networks and the reconstruction networks trained with dynamic routing algorithm;
	(c) The corresponding images reconstructed from the reconstruction networks trained in dynamic routing
	where the input capsules are obtained from our $l_2$-regularized routing algorithm without reconstruction networks.}
	\label{fig:reconstruction}
\end{figure*}

To verify the proposed methods, in this paper, we adopt the simplest capsule neural network architecture in \cite{CAP_NIPS2017_Hinton},
which is implemented with \texttt{tensorflow} \cite{DL_OSDI2016_TensorFlow}.
We conduct experiments on three datasets that include MNIST \cite{BM_MNIST_LeCun}, Fashion-MNIST \cite{BM_MNISTF_Han} and CIFAR-10 \cite{BM_CIFAR10_Alex}. 
Notations ``DR'', ``L1'' and ``L2''correspond to original dynamic routing \cite{CAP_NIPS2017_Hinton}, the proposed algorithm with $l_1$ regularization and $l_2$ regularization respectively. 
``x/FC'' denotes no fully connected reconstruction net is applied. 

\subsection{Neural Network Architecture}
In all the experiments, we adopt the simplest 3-layer capsule networks as used in \cite{CAP_NIPS2017_Hinton}, with one convolutional layer, one primary capsule layer and one output  layer.
Specifications are listed in \Cref{tab:arch}.
The first convolution layer is defined by 256 9$\times$9 kernels followed by two capsule layers with capsule vector dimensions of 8 and 16 respectively.
We use 32 primary capsules and 10 output capsules for the MNIST and Fashion-MNIST dataset and the primary capsule number is doubled when we are conducting experiments on CIFAR-10.
The reconstruction networks for MNIST dataset has 3 fully connected layers with neuron nodes of 512, 1024 and 784 respectively.
Reconstruction is not applied when training the network on CIFAR-10.
Each capsule layer is followed by the squash activation as in \Cref{eq:squash}.
We apply ReLU on the rest of the layers except the last layer in the reconstruction networks, which is equipped with sigmoid.

\subsection{Image Classification}

\begin{table}[tb!]
	\caption{Classification results of three benchmarks in terms of error rate (\%).}
	\label{tab:class}
	\centering
	\small
	\begin{tabular}{c|ccccc}
		\toprule
		Benchmarks    & DR \cite{CAP_NIPS2017_Hinton} & L2   & L1   & L2/FC & L1/FC \\ \hline 
		MNIST         & 0.34      & 0.35 & \textbf{0.32}  & 0.35      &0.44       \\
		Fashion-MNIST & 7.21      & 7.01 & 6.76 & \textbf{6.75}      &6.77       \\
		CIFAR-10      &15.3       & -    & - & 14.52    &  \textbf{14.04}       \\ \hline 
		Average       &7.62       & -    & - &7.21  &\textbf{7.08} \\ \bottomrule
	\end{tabular}
\end{table}

\begin{figure*}[tb!]
	\small
	\centering
	\subfloat[MNIST]{%{{{

\pgfplotsset{
    width =0.4\textwidth,
    height=0.28\textwidth,
    scaled x ticks=false
}
\begin{tikzpicture}[scale=1]
\begin{axis}[minor tick num=0,
xmin=15,
xmax=990,
ymax=2.0,
xticklabel style={/pgf/number format/.cd, fixed, fixed zerofill, precision=0, /tikz/.cd},
yticklabel style={/pgf/number format/.cd, fixed, fixed zerofill, precision=2, /tikz/.cd},
%scaled y ticks=base 10:0,
xlabel={Iteration Count ($\times$100)},
ylabel={Evaluation Error (\%)},
xlabel near ticks,
legend style={
  draw=none,
  at={(0.4,1.3)},
  anchor=north,
  legend columns=3,
}
]
\addplot +[line width=0.7pt] [color=mylight,  solid,  mark=none]  table [x={Iteration},  y={l2}, each nth point={3}]    {mnist.dat};
\addplot +[line width=0.7pt] [color=blue,   solid,  mark=none]  table [x={Iteration},  y={l2/fc}, each nth point={3}]   {mnist.dat};
\addplot +[line width=0.7pt] [color=black, solid,  mark=triangle]  table [x={Iteration},  y={dr}, each nth point={3}] {mnist.dat};
\addplot +[line width=0.7pt] [color=green,    solid,  mark=none]  table [x={Iteration},  y={l1}, each nth point={3}]  {mnist.dat};
\addplot +[line width=0.7pt] [color=red, solid,  mark=none]  table [x={Iteration},  y={l1/fc}, each nth point={3}] {mnist.dat};

\legend{L2, L2/FC, DR, L1, L1/FC}

\end{axis}
\end{tikzpicture}
%}}}
} \hspace{.3in}
	\subfloat[Fashion-MNIST]{%{{{

\pgfplotsset{
	width =0.4\textwidth,
	height=0.28\textwidth,
	scaled x ticks=false
}
\begin{tikzpicture}[scale=1]
\begin{axis}[minor tick num=0,
xmin=15,
xmax=2985,
ymax=12,
yticklabel style={/pgf/number format/.cd, fixed, fixed zerofill, precision=1, /tikz/.cd},
%scaled y ticks=base 10:0,
xlabel={Iteration Count ($\times$100)},
ylabel={Evaluation Error (\%)},
xlabel near ticks,
legend style={
	draw=none,
	at={(1.16,0.76)},
	anchor=north,
	legend columns=5,
}
]
\addplot +[line width=0.7pt] [color=mylight,  solid,  mark=none]  table [x={Iteration},  y={l2}, each nth point={10}]    {fmnist.dat};
\addplot +[line width=0.7pt] [color=blue,   solid,  mark=none]  table [x={Iteration},  y={l2/fc}, each nth point={10}]   {fmnist.dat};
\addplot +[line width=0.7pt] [color=green,    solid,  mark=none]  table [x={Iteration},  y={l1}, each nth point={10}]  {fmnist.dat};
\addplot +[line width=0.7pt] [color=red, solid,  mark=none]  table [x={Iteration},  y={l1/fc}, each nth point={10}] {fmnist.dat};
\addplot +[line width=0.7pt] [color=black, solid,  mark=triangle]  table [x={Iteration},  y={dr}, each nth point={10}] {fmnist.dat};
%\legend{L2, L2/FC, L1, L1/FC, DR}

\end{axis}
\end{tikzpicture}
%}}}
} \\
	\subfloat[CIFAR-10]{%{{{

\pgfplotsset{
	width =0.4\textwidth,
	height=0.28\textwidth,
	scaled x ticks=false
}
\begin{tikzpicture}[scale=1]
\begin{axis}[minor tick num=0,
xmin=15,
xmax=2325,
ymax=30,
yticklabel style={/pgf/number format/.cd, fixed, fixed zerofill, precision=1, /tikz/.cd},
%scaled y ticks=base 10:0,
xlabel={Iteration Count ($\times$100)},
ylabel={Evaluation Error (\%)},
xlabel near ticks,
legend style={
	draw=none,
	at={(1.16,0.76)},
	anchor=north,
	legend columns=5,
}
]
\addplot +[line width=0.7pt] [color=blue,   solid,  mark=none]  table [x={Iteration},  y={l2/fc}, each nth point={6}]   {cifar.dat};
\addplot +[line width=0.7pt] [color=red, solid,  mark=none]  table [x={Iteration},  y={l1/fc}, each nth point={6}] {cifar.dat};
\addplot +[line width=0.7pt] [color=black, solid,  mark=triangle]  table [x={Iteration},  y={dr}, each nth point={6}] {cifar.dat};
%\legend{L2, L2/FC, L1, L1/FC, DR}

\end{axis}
\end{tikzpicture}
%}}}
} \hspace{.3in}
	\subfloat[MultiMNIST]{%{{{

\pgfplotsset{
	width =0.4\textwidth,
	height=0.28\textwidth,
	scaled x ticks=false
}
\begin{tikzpicture}[scale=1]
\begin{axis}[minor tick num=0,
xmin=5,
xmax=180,
ymax=50,
yticklabel style={/pgf/number format/.cd, fixed, fixed zerofill, precision=1, /tikz/.cd},
%scaled y ticks=base 10:0,
xlabel={Iteration Count ($\times$100)},
ylabel={Evaluation Error (\%)},
xlabel near ticks,
legend style={
	draw=none,
	at={(1.16,0.76)},
	anchor=north,
	legend columns=5,
}
]
\addplot +[line width=0.7pt] [color=blue,   solid,  mark=none]  table [x={Iteration},  y={l2/fc}, each nth point={1}]   {mmnist.dat};
\addplot +[line width=0.7pt] [color=red, solid,  mark=none]  table [x={Iteration},  y={l1/fc}, each nth point={1}] {mmnist.dat};
\addplot +[line width=0.7pt] [color=black, solid,  mark=triangle]  table [x={Iteration},  y={dr}, each nth point={1}] {mmnist.dat};
%\legend{L2, L2/FC, L1, L1/FC, DR}

\end{axis}
\end{tikzpicture}
%}}}
}
	\caption{Visualization of the convergence on different routing algorithms. 
		(a)--(d) are regular training curves on MNIST, Fashion-MNIST, CIFAR-10 and MultiMNIST, respectively.}
	\label{fig:train}
\end{figure*}
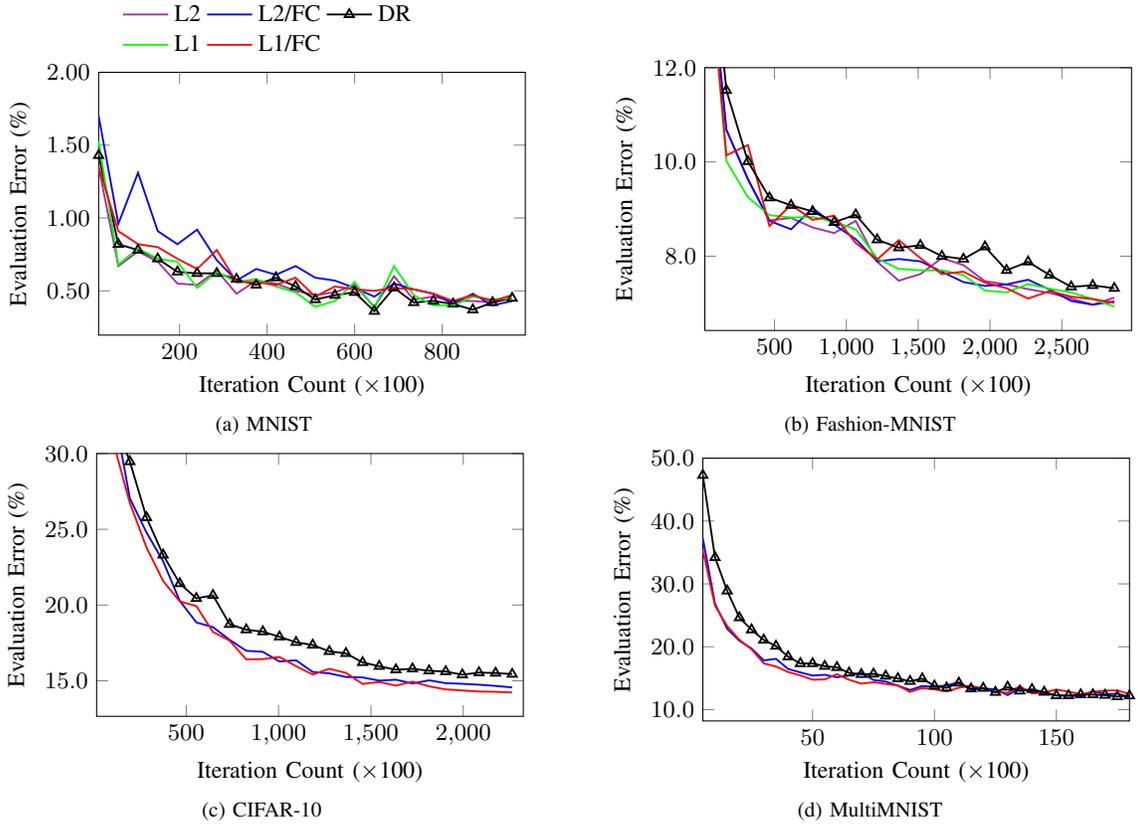

In the first experiment, we compare the classification results with \cite{CAP_NIPS2017_Hinton} on three benchmarks discussed above as shown in \Cref{tab:class}.
%Column ``\textbf{DR}'' corresponds to the result of single round dynamic routing per iteration in \cite{CAP_NIPS2017_Hinton} with exactly the same settings;
%Columns ``\textbf{L2}'' and ``\textbf{L1}'' list the results of the proposed routing methods with $l_2$ and $l_1$ regularization respectively;
%Columns ``\textbf{L2/FC}'' and ``\textbf{L1/FC}'' are the corresponding to results of our methods without reconstruction networks as a regularization term.%, which we will show later the reconstruction regularization limits the classification solution space.

For the MNIST dataset, %we train the model with a batch size of 128 and a starting learning rate of 0.001 that decays by 0.96 per 1000 iterations.
we observe that the margin loss drops fast even at early training stage when fully connected reconstruction networks are removed. 
Therefore we pick a smaller batch size (i.e.~32) when training the capsule networks without reconstruction networks.
The learning rate decays by 0.5 every 1000 iterations.
Because deep learning models can easily achieves above 99.0\% accuracy on MNIST, it is hard to have further significant improvements.
Here we only show similar classification results compared to dynamic routing. 

We train the Fashion-MNIST dataset with a batch size of 128 and a starting learning rate of 0.001 that decays by 0.96 every 1000 iterations.
%We pick $\lambda=0.00001$ for both $l_2$ and $l_1$ regularized routing algorithms, while the configuration of dynamic routing is kept as it originally is.
We also train the network without reconstruction layers and keep everything else unchanged.
The test error rate drops from 7.21\% to 7.01\% and 6.76\% when using our routing algorithms with $l_2$ and $l_1$ regularization respectively.
It can also be seen that the classification error further drops when the reconstruction networks are removed especially on the $l_2$ regularized algorithm.

The CIFAR-10 model is trained on a single capsule networks (without any model ensemble) using the architecture specified in \Cref{tab:arch} where the number of primary capsules is doubled and reconstruction networks are removed as in \cite{CAP_NIPS2017_Hinton}.
%The network is trained with a batch size of 128 and an initial learning rate of 0.001 which is decayed by 0.96 every 2000 iterations.
We also replace the classic ReLU with LeakyReLU \cite{DL_ICML2013_Maas} during training that shows better performance.
As shown in \Cref{tab:class}, our routing methods with $l_2$ and $l_1$ regularization reduce the single model classification error rate by 0.78\% and 1.26\% respectively compared to dynamic routing.

\subsubsection{Discussion of FC-Regularization} 
\Cref{tab:class} shows smaller classification error when the neural network is trained without fully-connected reconstruction nets.
One explanation is that the training set is not necessarily covering the whole data space.
In this experiment, we evaluate the trained neural networks on Fashion-MNIST with and without reconstruction regularization.
We feed correctly activated output capsule vectors associated to the model without reconstruction networks into the reconstruction networks and obtain the reconstructed images as shown in \Cref{fig:reconstruction}.
It can be seen that a fraction of those images are not correspondingly reconstructed but correctly classified,
which agrees with \Cref{tab:class}.
For the classification purpose only, removing the reconstruction networks can also improve the training efficiency by dropping redundant trainable variables.

\subsubsection{Segmenting Overlapped Digits}
We also conduct experiments to show our routing solution attains the ability to recognize overlapped digits.
In this experiment, we adopt the MultiMNIST dataset as used in \cite{CAP_NIPS2017_Hinton},
where two digits from different classes are overlapped together with at most 4 pixels shift in each direction that forms into 36$\times$36 images.
The MultiMNIST dataset contains 60 million training samples and 10 million testing samples.
We train the capsule nets using $l_2$ and $l_1$ regularized routing algorithms respectively.
The initial learning rate is set 0.001 that decays by 0.96 every 20000 iterations.
We also set a larger regularization coefficient with $\lambda=0.001$ to avoid over-fitting.
Because the MultiMNIST training set is extremely large we train the neural networks for 200000 steps for both ``DR'' and our routing algorithms,
when we achieved an evaluation error of 7.54\% compared to 7.47\% of dynamic routing.
It should be noted that although the evaluation errors are similar for both methods, the training speed is relatively faster than dynamic routing, as discussed in the following section.

\subsection{Convergence of the Routing Algorithm}

\begin{figure}[tb!]
	\small
	\centering
	\pgfplotsset{scaled y ticks=false}
\begin{tikzpicture}
\pgfplotsset{
	width =0.45\textwidth,
	height=0.25\textwidth,
	every axis plot/.append style = {font = \tiny},
	/pgfplots/bar cycle list/.style={/pgfplots/cycle list={%
{black,fill=myblue},
		}
	},
}
\begin{axis}[
ybar=1pt,
bar width=7pt,
legend style={at={(0.5,1.2)},
	anchor=north,legend columns=-1},
area legend,
ylabel={Runtime ($s$)},
symbolic x coords={DR, L1, L2, L1/FC, L2/FC},
xtick=data,
ymin=0,
ytick={5,10,15,20},
ylabel near ticks,
 x tick label style={rotate=0,font=\small},
%nodes near coords,
%nodes near coords align={vertical},
]
\addplot  coordinates {(DR,18.79) (L1,14.98) (L2,15.61)(L1/FC,11.92) (L2/FC,11.10)};

%\legend{SPIE'15,ICCAD'16,Ours}
\end{axis}
\end{tikzpicture}
	\caption{Approximate training time per 100 steps.}
	\label{fig:time}
\end{figure}
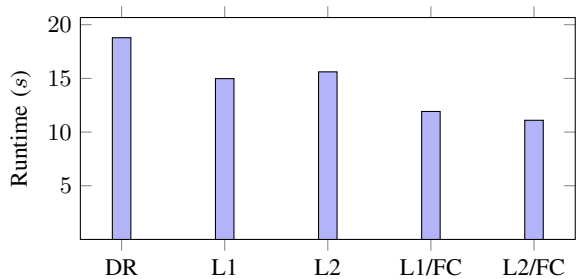

In support of the proposed routing algorithm, we visualize the evaluation performance of the capsule networks along with the training procedure in \Cref{fig:train}.
All models show similar convergence curves on MNIST dataset that reach an evaluation error under 0.4\%.
For the more challenging Fashjon-MNIST and CIFAR-10, all regularized routing algorithms discussed in this work exhibit faster and better convergence in terms of evaluation error and hence demonstrates the effectiveness and the efficiency of our methods.
Because the MultiMNIST dataset is extremely large, we only visualize the training behavior in 20000 steps.
We can observe that the evaluation error drops much faster than dynamic routing at early training steps, which is consistent with the discussion about Assumption~\ref{asm:2}.
Our algorithm also shows an advantage in terms of training runtime that each step can save at least 20\% runtime compared to dynamic routing (DR), as shown in \Cref{fig:time}.

%\subsection{On the Effectiveness of Regularization}

\section{Conclusion}
\label{sec:conclu}

The basis of capsule neural networks and associated routing algorithms are studied in this paper,
based on which a new objective on determining routing coefficients between capsules are established.
An algorithm targeting at the new routing objective is proposed to achieve faster model convergence and competitive classification results,
compared to the baseline results achieved by dynamic routing algorithm on the same capsule network architecture.
We also discuss the effectiveness of fully connected reconstruction networks in support of the classification results and visualized counterexamples.
Additional researches on development efficient capsule network architecture and hyper-parameter exploration to compete with state-of-the-art solutions on larger datasets (e.g.~ImageNet \cite{BENCHMARK_IMGNET_jia}) will be conducted in our future work.

{
    \bibliographystyle{IEEEtran}
    \bibliography{./ref/Top,./ref/DFM,./ref/HSD,./ref/Software,./ref/LEARN,./ref/DL,./ref/MTRX,./ref/additional}
}

\end{document}